\documentclass[letterpaper]{article} 
\usepackage{aaai25}  
\usepackage{times}  
\usepackage{helvet}  
\usepackage{courier}  
\usepackage[hyphens]{url}  
\usepackage{graphicx} 
\usepackage{natbib}  
\usepackage{caption} 
\usepackage{algorithm}
\usepackage{algorithmic}

\usepackage{multirow}

\usepackage{amssymb}
\usepackage{arydshln}
\usepackage{booktabs}
\usepackage{footnote}
\usepackage{amsthm}
\usepackage{amsmath}
\usepackage{mathrsfs}
\usepackage{graphicx}

\usepackage{colortbl}
\usepackage{bbding}
\usepackage{float}
\usepackage{utfsym}

\usepackage[misc]{ifsym}

\usepackage{newfloat}
\usepackage{listings}
\DeclareCaptionStyle{ruled}{labelfont=normalfont,labelsep=colon,strut=off} 
\floatstyle{ruled}
\newfloat{listing}{tb}{lst}{}
\floatname{listing}{Listing}
\title{ARNet: Self-Supervised FG-SBIR with Unified Sample Feature Alignment \\ and Multi-Scale Token Recycling}
\author {
    Jianan Jiang$^\dag$$^\$$,
    Hao Tang$^\ddag$$^*$,
    Zhilin Jiang$^\dag$,
    Weiren Yu$^\S$,
    Di Wu$^\dag$$^\$$$^*$
}
\affiliations {
    $^\dag$Hunan University,
    $^\ddag$Peking University,
    $^\S$University of Warwick,
    $^\$$ExponentiAI Innovation\\
    \{jiangjn22, zhilin, dwu\}@hnu.edu.cn, haotang@pku.edu.cn, weiren.yu@warwick.ac.uk
}

\usepackage{bibentry}

\begin{document}

\maketitle

\renewcommand{\thefootnote}{\fnsymbol{footnote}}
\footnotetext[1]{Corresponding Authors.}

\begin{abstract}
\label{abstract}
Fine-Grained Sketch-Based Image Retrieval (FG-SBIR) aims to minimize the distance between sketches and corresponding images in the embedding space. However, scalability is hindered by the growing complexity of solutions, mainly due to the abstract nature of fine-grained sketches. In this paper, we propose an effective approach to narrow the gap between the two domains. It mainly facilitates unified mutual information sharing both intra- and inter-samples, rather than treating them as a single feature alignment problem between modalities. Specifically, our approach includes: (i)~Employing dual weight-sharing networks to optimize alignment within the sketch and image domain, which also effectively mitigates model learning saturation issues. (ii)~Introducing an objective optimization function based on contrastive loss to enhance the model's ability to align features in both intra- and inter-samples. (iii)~Presenting a self-supervised \textit{Multi-Scale Token Recycling (MSTR) Module} featured by recycling discarded patch tokens in multi-scale features, further enhancing representation capability and retrieval performance. Our framework achieves excellent results on CNN- and ViT-based backbones. Extensive experiments demonstrate its superiority over existing methods. We also introduce Cloths-V1, the first professional fashion sketch-image dataset, utilized to validate our method and will be beneficial for other applications. Our code, new datasets, and pre-trained model are available at \url{https://github.com/ExponentiAI/ARNet}.

\end{abstract}

%

\section{Introduction}

\begin{figure}[t]
\centering
\includegraphics[width=\linewidth]{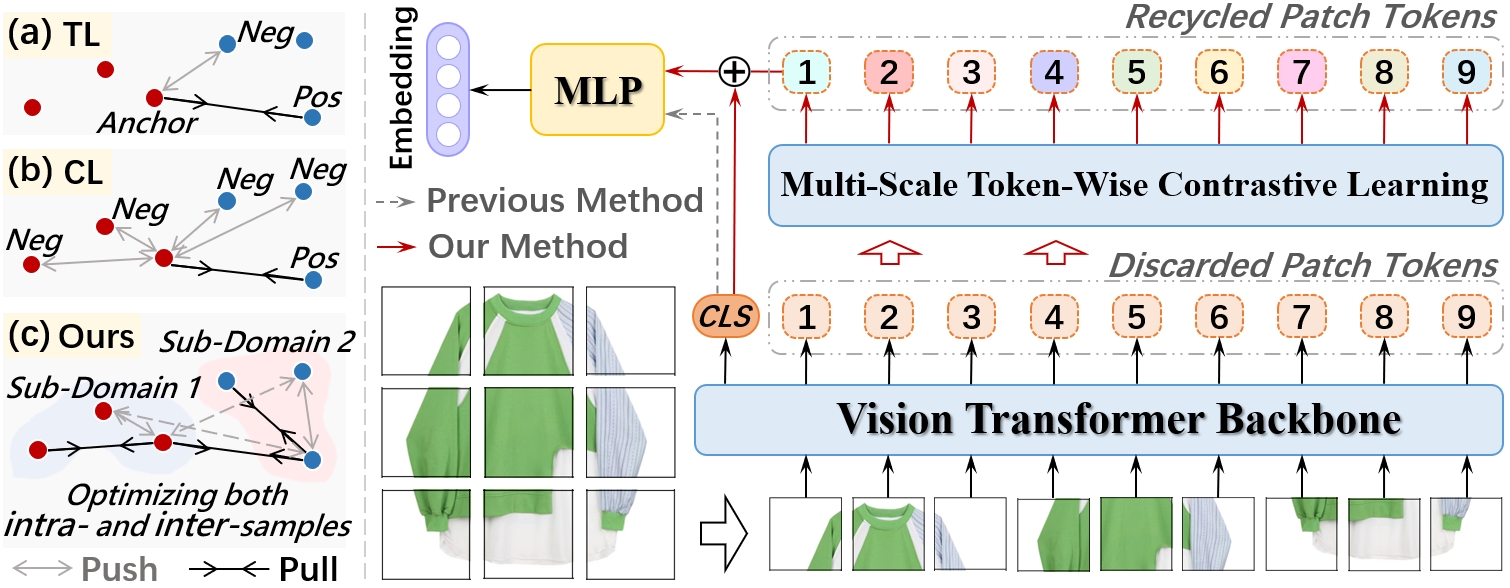}
\caption{$\mathtt{[Left]}$ Comparison of our objective function with traditional methods based on Triplet Loss (TL) and Contrastive Loss (CL). $\mathtt{[Right]}$ Comparison of our method with previous methods for handling Patch Tokens in ViTs. }
\label{fig:loss}
\end{figure}

Starting with~\cite{yu2016sketch}, the triplet loss has gained prominence in the field of Fine-Grained Sketch-Based Image Retrieval (FG-SBIR). It encourages proximity among similar samples (Pos) in the embedding space while pushing different samples (Neg) farther apart, as shown in Fig.~\ref{fig:loss}~(a), thereby achieving effective separation of the samples. However, the selection of suitable triplets necessitates further refinement to achieve optimal performance. Moreover, in some cases, the triplet loss might not provide sufficient gradient signals, thereby affecting the model optimization process. To address these challenges, methods such as reinforcement learning~\cite{muhammad2018learning,bhunia2020sketch,bhunia2022sketching}, meta-learning~\cite{sain2021stylemeup,bhunia2022adaptive}, and out-of-sample learning~\cite{sain2023exploiting} have emerged. These approaches aim to refine the model's retrieval capabilities. However, such strategies may introduce instability during model training and pose challenges in tuning hyperparameters, such as gradient vanishing and hyperparameter sensitivity in~\cite{sain2021stylemeup, bhunia2021more}. These challenges collectively compromise the overall model scalability, making it challenging for researchers in the community to further expand their research. Furthermore, several methods~\cite{sain2020cross,yang2021sketchaa,pang2020solving} divide input sketches into strokes or layers, adding additional complexity. These might be less compatible with other datasets and less adaptable for broader exploration. In summary, triplet loss has achieved widespread adoption and spurred the development of techniques centered on its selection and optimization. However, this advancement has come at the cost of modeling complexity and a compromise in stability.

The introduction of Vision Transformer (ViT)~\cite{dosovitskiy2020image} has introduced a new architectural backbone and powerful performance in the field of computer vision. However, existing methods~\cite{Chaudhuri2022XModalViT, sain2023exploiting} typically utilize the \textit{[CLS] Token} as feature output, discarding the features of other \textit{Patch Tokens} as shown in Fig.~\ref{fig:loss}, or employing common aggregation methods such as AvgPool or MaxPool, etc. We found that there are differences between these discarded \textit{Patch Tokens}, which may contain potential useful information. Therefore, reasonable recycling of useful features from these discarded tokens will improve the representation ability of the model.

In this study, we introduce the ARNet (Feature \textbf{A}lignment and Token \textbf{R}ecycling). We experimentally discovered that a straightforward methodology can achieve favorable results and enhance the model's versatility by switching any backbone. Additionally, recycling token features can improve the model's feature representation capability. Specifically, we simultaneously optimize feature distribution alignment intra- and inter-samples through structural design and loss function configuration. Our initial experiments revealed that single-branch encoders can tend to train saturation. Therefore, we introduced a dual weight-sharing network structure to enhance feature consistency within the sketch and image domains, ensuring effective learning of key fine-grained features. We also proposed a new objective optimization method for self-supervised training, promoting optimization of features in both intra- and inter-samples, as shown in Fig.~\ref{fig:loss} (c), which differs from traditional Triplet Loss (a) and Contrastive Loss (b). Furthermore, we introduced the plug-and-play \textit{Multi-Scale Token Recycling (MSTR) Module} to recycle discarded \textit{Patch Tokens} and used Multi-Scale Token-Wise Contrastive Learning to filter out similar features between tokens while retaining unique features. This approach enhances feature learning from individual samples, thereby improving the model's representation ability.

In summary, our contributions are summarized as follows:

\begin{itemize}
\item We propose a novel self-supervised FG-SBIR framework straightforwardly with the dual weight-sharing networks to enhance feature alignment within the samples domain and the objective optimization function to optimize intra- and inter-sample feature distributions concurrently.
\item We introduce a plug-and-play module for ViT, the MSTR, which leverages the discarded patch tokens recycling and contrastive learning among tokens with multi-scale features to achieve better feature representation and enhanced performance.
\item Extensive experiments verify the effectiveness of each component and the superior performance of our method on FG-SBIR over other state-of-the-art methods.
\item Additionally, we introduce a new dataset named Clothes-V1, filling the gap in professional fashion clothing datasets in this field. Its multi-level quality can be valuable for other computer vision tasks as well.
\end{itemize}

\section{Related Work}
\subsubsection{FG-SBIR.}
Fine-grained sketch-based image retrieval (FG-SBIR) involves retrieving images in a specific category based on a given sketch query. The evolution of FG-SBIR, from the introduction of triplet loss~\cite{yu2016sketch} to advancements in model learning capabilities through attention mechanisms~\cite{song2017deep} and recent approaches such as reinforcement learning~\cite{muhammad2018learning,bhunia2020sketch,bhunia2022sketching}, meta-learning~\cite{sain2021stylemeup,bhunia2022adaptive}, and sketch layered feature representation~\cite{bhunia2020sketch,sain2020cross,yang2021sketchaa}, has been rapid. However, these approaches are based on triplet loss and improve model performance using intricate methodologies. In contrast, our approach takes a different perspective by designing a simple and universal framework. This is suitable for any backbone, yielding improved retrieval results without a complex methodology design.

\subsubsection{Contrastive Learning.}
Contrastive learning improves model representational ability by maximizing positive sample similarity and minimizing negative sample similarity. 
In earlier works,~\citet{wu2018unsupervised} proposed the memory bank to store data features. Since then, methods like~\cite{cvpr19unsupervised, tian2020contrastive, chen2020simple,duan2024mining} used data augmentation, learned from multiple views as input or larger pos-neg sample pairs, etc.~\citet{he2020momentum} proposed the momentum encoder to address sample selection bias.~\citet{caron2020unsupervised} used a clustering algorithm to reduce computational costs. Recently,~\citet{wang2021exploring, tang2020edge, tang2023edge} proposed pixel contrast for image segmentation or synthesis. These methods drive unsupervised representation learning to achieve superior performance. Similarly, we can leverage the concept of contrastive learning to FG-SBIR for better retrieval performance.

\subsubsection{Vision Transformer.}
Vision Transformer (ViT)~\cite{dosovitskiy2020image} uses a self-attention mechanism like~\cite{vaswani2017attention}, leading to a more effective way to model global features compared to traditional CNNs. Based on this, methods such as~\cite{han2021tnt, thiry2024towards, liu2021Swin, wang2021pyramid} proposed smaller patches, the shift-window mechanism, progressive shrinking pyramid, etc. to further enhance the capability of ViTs. However, only a few studies~\cite{Chaudhuri2022XModalViT, sain2023exploiting} have explored ViT in the field of FG-SBIR. Moreover, these methods choose the \textit{[CLS] Token} as feature output and discard other \textit{Patch Tokens}, which will also result in discarding some potentially useful features. Therefore, we introduce the ViT into FG-SBIR and further improve model performance by recycling the discarded features.

\section{Methodology}
\label{section:mothod}

\subsection{How to Keep Learning?}

\begin{figure}[t]
\centering
\includegraphics[width=\linewidth]{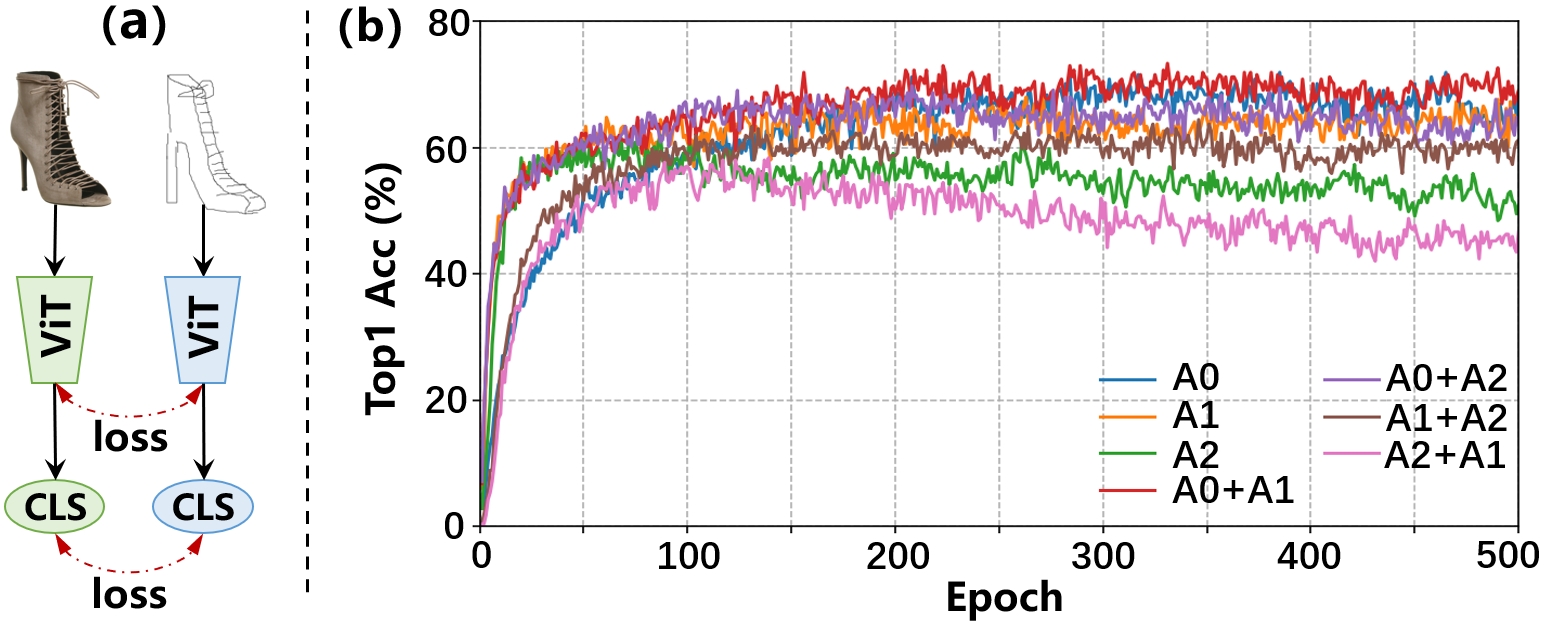}
\caption{(a) A initial network for FG-SBIR. (b) Visualization of the training process with different augmentations.}
\label{fig:net}
\end{figure}

\begin{figure*}[t]
\centering
\includegraphics[width=0.9\linewidth]{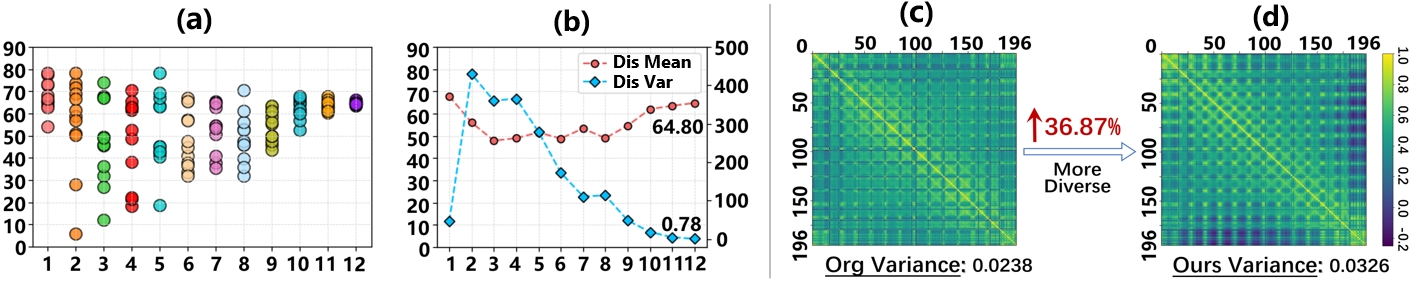}
\caption{(a)(b) The average attention distance of ViT backbone in multi-heads (dots) w.r.t each layer. (c)(d) The comparison of similarity and variance between discarded 196 \textit{Patch Tokens} in the output of the ViT backbone.}
\label{fig:distance}
\vspace{-2mm}
\end{figure*}

We begin with an initial structure as shown in Fig.~\ref{fig:net} (a), using the prevalent pre-trained ViT-B/16~\cite{dosovitskiy2020image} as our backbone and modifying the dimensions of the classification layer. However, we encountered the first challenge: the model's performance exhibited saturation around the 150-epoch mark, and subsequent training iterations yielded negligible enhancements, as illustrated by the A0 curve in Fig.~\ref{fig:net} (b)~\footnote{A0: no augmentation. A1: image scaling, rotation, and cropping operations. A2: similar to Moco~\cite{he2020momentum}.}. We tried experimenting with conventional solutions like data augmentation, output feature normalization, etc. Unfortunately, these methods did not show improvements. In terms of data augmentation, methods like A1 and A2 appeared to have no benefit in the generalization capability or even drive the model into overfitting territory, as shown in Fig.~\ref{fig:net} (b) by the A1 and A2 curves.

Thus, the question of whether the model's ability to learn better feature representations from input samples was impeded by its single-network architecture. Inspired by~\cite{bromley1993signature}, we surpass this limitation by introducing the dual weight-sharing networks. This change in approach enabled the model to optimize the feature distribution of both intra- and inter-samples, yielding more representative and discriminative features. Furthermore, through an alternative preprocessing approach for inputs, we have a noteworthy observation: the inclusion of A0 is beneficial to the weight-sharing network and consistently yields better outcomes. Intriguingly, the elaborate data augmentation strategy A2 is incompatible with the nuances of fine-grained datasets, dampening the model's generalization abilities. The A0+A1 curve in Fig.~\ref{fig:net} (b) demonstrates the model's enduring generalization capacity within the 500 epochs with the absence of pronounced feature degradation.

\subsection{Are Discarded Patch Tokens Useless?}
In previous studies~\cite{dosovitskiy2020image}, the features at the \textit{[CLS] Token} were chosen as the features extracted by ViT, and the features at other \textit{Patch Tokens} were discarded. So are these discarded features really useless?

The Multi-Head Self-Attention module is the core component of the ViT, aiding in the analysis of information and attention distribution learned by the model. For the $n$-th head, the attention matrix can be expressed as:
\begin{equation}
\mathbf{Att}^n = \mathtt{softmax}\left(\mathbf{Q}^n (\mathbf{K}^n)^T/ \sqrt{d_{\mathbf{K}^n}}\right) \cdot \mathbf{V}^n,
\end{equation}
where $n\in\{1, \dots, N\}$, $N$ is the number of multi-heads with each head computing the attention independently, followed by concatenation and projection to form the final output. $\mathbf{Q}$ (query), $\mathbf{K}$ (key), and $\mathbf{V}$ (value) are linear projections of the input feature, obtained by applying three learnable weight matrices $W^Q$, $W^K$, and $W^V$, respectively.

As expressed in Eq.~\eqref{avgvar}, we calculate the mean and variance of the average attention of each head. Here, $\textbf{Att}_{ij}$ is the attention weight of the $i$-th token to the $j$-th token, $|i - j|$ is the distance between the $i$-th token and the $j$-th token, and $\mu_i$ is the average attention distance of the $i$-th token. $M$ is the number of patch tokens. $\textbf{Mean}$ and $\textbf{Var}$ are the calculated mean and variance average attention distance, respectively.
\begin{equation}
\label{avgvar}
\begin{split}
\mu_i = \sum_{j=1}^M &\mathbf{Att}_{ij}^n |i - j|, \quad
\textbf{Mean}^n = \frac{1}{M} \sum_{i=1}^M \mu_i, \\
\textbf{Var}^n &= \frac{1}{M} \sum_{i=1}^M \sum_{j=1}^M (\mathbf{Att}_{ij}^n |i - j| - \mu_i)^2.
\end{split}
\end{equation}

As illustrated in Fig.~\ref{fig:distance}, we compute the average attention distance of multi-heads attention for each layer within the ViT backbone, along with the \textit{Mean} and \textit{Variance} of attention distance for each layer. A lower \textit{Mean} indicates a model preference for local features, while a lower \textit{Variance} indicates more consistent feature representations. We observe that the model's attention map tends to be consistent as the number of model layers increases (Fig.~\ref{fig:distance}~(a)), but it is not completely consistent (Fig.~\ref{fig:distance}~(b)). There are always some differences between the \textit{Patch Tokens} in each head (Fig.~\ref{fig:distance}~(c)), and these differences likely contain useful semantic information. By recycling these discarded \textit{Patch Tokens}, we can extract the unique features contained in each token and integrate them into the \textit{[CLS] Token}, achieving better feature representation. Fig.~\ref{fig:distance}~(c) and (d) present the feature similarity graphs with and without the \textit{Patch Tokens} recycling, respectively. Our method increases feature differences by 36.87\%, thereby ensuring that the unique features of each token are retained and utilized effectively.

\begin{figure*}
\centering
\includegraphics[width=\linewidth]{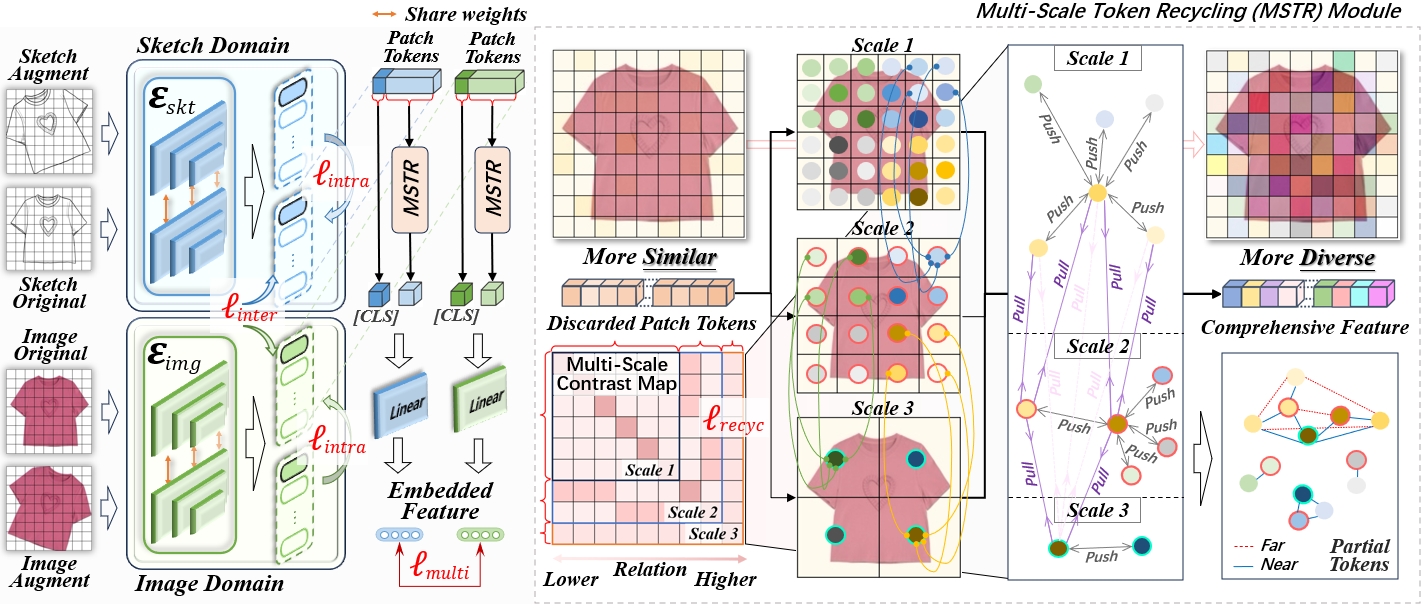}
\caption{An overview of our method, which enhanced feature alignment from both intra- and inter-sample perspectives. We propose the \textit{MSTR Module} to further improve the Encoder’s feature representation ability by recycling discarded \textit{Patch Tokens}.}
\label{fig:framework}
\vspace{-3mm}
\end{figure*}

\subsection{Feature Alignment and Token Recycling}
We present the framework of our method in Fig.~\ref{fig:framework}, which consists of three main components: (i) The \textit{Basic Model} that achieves high retrieval performance with a straightforward structure by feature alignment. (ii) The \textit{Multi-Scale Token Recycling (MSTR)} Module designed to recycle potentially valuable features from discarded \textit{Patch Tokens}. (iii) The \textit{Loss Function} utilized for optimizing model training.

\subsubsection{Basic Model.}
We begin by randomly loading sketch-image sample pairs from the dataset and performing data augmentation. The augmented and original samples are then input into the weight-sharing network $\mathcal{E}$ as previously described. Finally, we utilize a linear layer to map the \textit{[CLS] Token} of sketch and image features to a joint embedding space. Specifically, denoting the sketches and images within our dataset as $\mathcal{S}$ and $\mathcal{I}$ respectively, with the number of samples in each minibatch designated as \textit{m}, we succinctly represent our data-loading process as $\mathcal{B}_m \Rightarrow \{s_j \in \mathcal{S}, i_j \in \mathcal{I} \mid s_j \text{ and } i_j \in \mathcal{R}^{224 \times 224 \times 3}\}$, and $j = 1, 2, \dots, m$. Initiating the feature extraction process, we begin by applying data augmentation to the input samples $\mathcal{B}_m$, generating the sample pairs denoted as $\mathcal{B}'_m \Rightarrow \{(s_j, s'_j, i_j, i'_j) \mid s_j \in \mathcal{S}, i_j \in \mathcal{I}, s'_j \in \mathcal{S} \text{ and } i'_j \in \mathcal{I} \text{ are augmented samples}\}$. Subsequently, the augmented samples undergo patch encoding and positional encoding before being fed to the encoders $\mathcal{E}_{skt}$ and $\mathcal{E}_{img}$, obtaining the sketch features $\mathcal{F}_{skt} \Rightarrow \{f_{skt} \leftarrow \mathcal{E}_{skt}(s), f'_{skt} \leftarrow \mathcal{E}_{skt}(s')\}$ and image features $\mathcal{F}_{img} \Rightarrow \{f_{img} \leftarrow \mathcal{E}_{img}(i), f'_{img} \leftarrow \mathcal{E}_{img}(i')\}$ for the input samples $\mathcal{B}_m$. Finally, the resulting features $\mathcal{F}_{skt}$ and $\mathcal{F}_{img}$ are channeled through a linear layer, mapping them into the $\mathcal{R}^{1 \times 512}$ dimension within a joint embedding space. This process yields the embedded sketch features $\mathcal{F}_{skt}^{out}$ and image features $\mathcal{F}_{img}^{out}$ for retrieval in a straightforward way.

\subsubsection{MSTR Module.}
Based on the previous analysis and the \textit{Basic Model}, we further propose the \textit{Multi-Scale Token Recycling (MSTR) Module}, which can achieve a better feature representation of samples by recycling potentially useful features from discarded \textit{Patch Tokens}. Specifically, for the feature 
$\mathcal{F}$ obtained by the ViT backbone, which contains the features of retained \textit{[CLS] Token} $\mathcal{F}_{CLS}$ and discarded \textit{Patch Tokens} $\mathcal{F}_{PT}$. We use the \textit{MSTR Module} to extract potentially useful features from the $\mathcal{F}_{PT}$, then output and add them to the $\mathcal{F}_{CLS}$. Other operations remain consistent with the \textit{Basic Model}, making \textit{MSTR} a plug-and-play module.

For the input feature $\mathcal{F}_{PT}$, we recognize that after the self-attention mechanism of ViT, the features among these tokens become $\mathtt{More\ Similar}$. To uncover potentially useful information, we leverage the fact that each token originates from a different position in the image, thus containing unique semantic information. Our goal is to make the final $\mathcal{F}_{PT}$ $\mathtt{More\ Diverse}$, ensuring each token retains unique and useful features, thereby improving the overall feature representation. We first reduce the dimension of $\mathcal{F}_{PT}$ to obtain three scale features, providing a richer multi-level feature representation $\mathcal{F}_{ML}$. We then use two Transformer layers~\cite{vaswani2017attention} to sequentially encode the $\mathcal{F}_{ML}$. Finally, the resulting embedded feature is dimensionally reduced and added with $\mathcal{F}_{CLS}$.

During the model training process, we reduce the dimensionality of the encoded $\mathcal{F}_{ML}$ to filter out some information and then concatenate them to obtain the feature $\mathcal{F}_{ML}^{map}$. Furthermore, we constructed a \textit{Multi-Scale Contrast Map} ($\mathtt{MAP}$) to guide the learning of the \textit{MSTR Module}. For tokens within the same scale, we maximize the distance between them to highlight their differences. For tokens between different scales, we maximize the distance between patches at different positions and minimize the distance between patches at the same or adjacent positions, ensuring that we capture unique features while not ignoring the inherent similarities among tokens. This process can be represented as:
\begin{equation}
\mathcal{L_{M}} = 
\begin{matrix}
\underbrace{\frac{1}{n^2} \sum_{i=1}^{n}\sum_{j=1}^{n}\left((\mathcal{F}_{ML}^{map} \cdot {\mathcal{F}_{ML}^{map}}^T)_{ij} - \mathtt{MAP}_{ij}\right)^2}.
\\ \ell_{recyc}
\end{matrix}
\end{equation}

\subsection{Loss Function}
\label{section:loss}
Inspired by~\cite{chen2020simple}, we introduce contrastive loss as our basic objective function:
\begin{equation}
\label{contrastive_cross}
\begin{split}
\mathcal{L}(\mathcal{F}) = -\frac{1}{N}\sum_{i=1}^N\log\frac{\sum_{z=1}^{{\frac{L}{N}}-1}\exp((\mathcal{F}_i \cdot \mathcal{F}_{N \cdot z  + i}^\top)/{\tau})}{\sum_{j=1}^{L} \mathbb{I}_{[i \neq j]} \exp((\mathcal{F}_{i} \cdot \mathcal{F}_{j}^\top)/{\tau})},
\end{split}
\end{equation} 
where $\mathcal{F}$ denotes an input feature vector derived from $n$ sample vectors and concatenate $\mathtt{Cat}$ along the batch size dimension, its construction follows $\mathcal{F} = \mathtt{Cat}(\mathcal{F}_1, \mathcal{F}_2, \dots, \mathcal{F}_n)$. $N$ is the batch size, $L$ is the length of the concatenated feature vector, and $\tau$ is a temperature parameter. The indicator function $\mathbb{I}{[i \neq j]} \in \{0, 1\}$ equals 1 when $i \neq j$. $(\cdot)$ is the dot product to obtain the similarity matrix between two vectors.

\subsubsection{Basic Model Optimize.}
The complete loss function of our Basic Model (ARNet Basic) comprised of: (i) \textit{Multi-Modal Loss} $\ell_{multi}$ minimizes the distance between the sketch and its corresponding image within the joint embedding space. (ii) \textit{Inter-Sample Loss} $\ell_{inter}$ encourages the similarity between the sketch features and image features output by their Encoders $\mathcal{E}$, which can further optimize their ultimate distribution within their joint embedding space. (iii) \textit{Intra-Sample Loss} $\ell_{intra}$ learns the distinctive features inherent to the original images and their augmented counterparts in a self-supervised way, enhancing the capability of feature representation. In general, it is represented as follows:
\begin{equation}
\begin{split}
\mathcal{L_{B}} = \ &
\begin{matrix}
\underbrace{\mathcal{L}(\mathcal{F}_{skt}^{out}, \mathcal{F}_{img}^{out})}
\\ \ell_{multi}
\end{matrix} \
\begin{matrix}
+ \underbrace{\mathcal{L}(f_{skt}, f_{img})}
\\ \ell_{inter}
\end{matrix} \\ &
\begin{matrix}
+ \underbrace{\mathcal{L}(f_{skt}, f'_{skt}) + \mathcal{L}(f_{img}, f'_{img})}.
\\ \ell_{intra}
\end{matrix}
\end{split}
\end{equation}

\subsubsection{Full Model Optimize.}
Based on the Basic Model, we incorporate the MSTR Module to achieve better performance. As a result, the loss function of our full model (ARNet Full) can be represented as:
\begin{equation}
\mathcal{L_{+}} = \mathcal{L_{B}} + \mathcal{L_{M}}.
\end{equation}

In this way, our approach can facilitate the model to simultaneously focus on information both intra- and inter-samples with multi-perspectives, while incorporating recycled potentially valuable features. This results in comprehensive feature representations and enhances overall retrieval performance.

\section{Experiments and Results}

\begin{figure}[t]
\centering
\includegraphics[width=\linewidth]{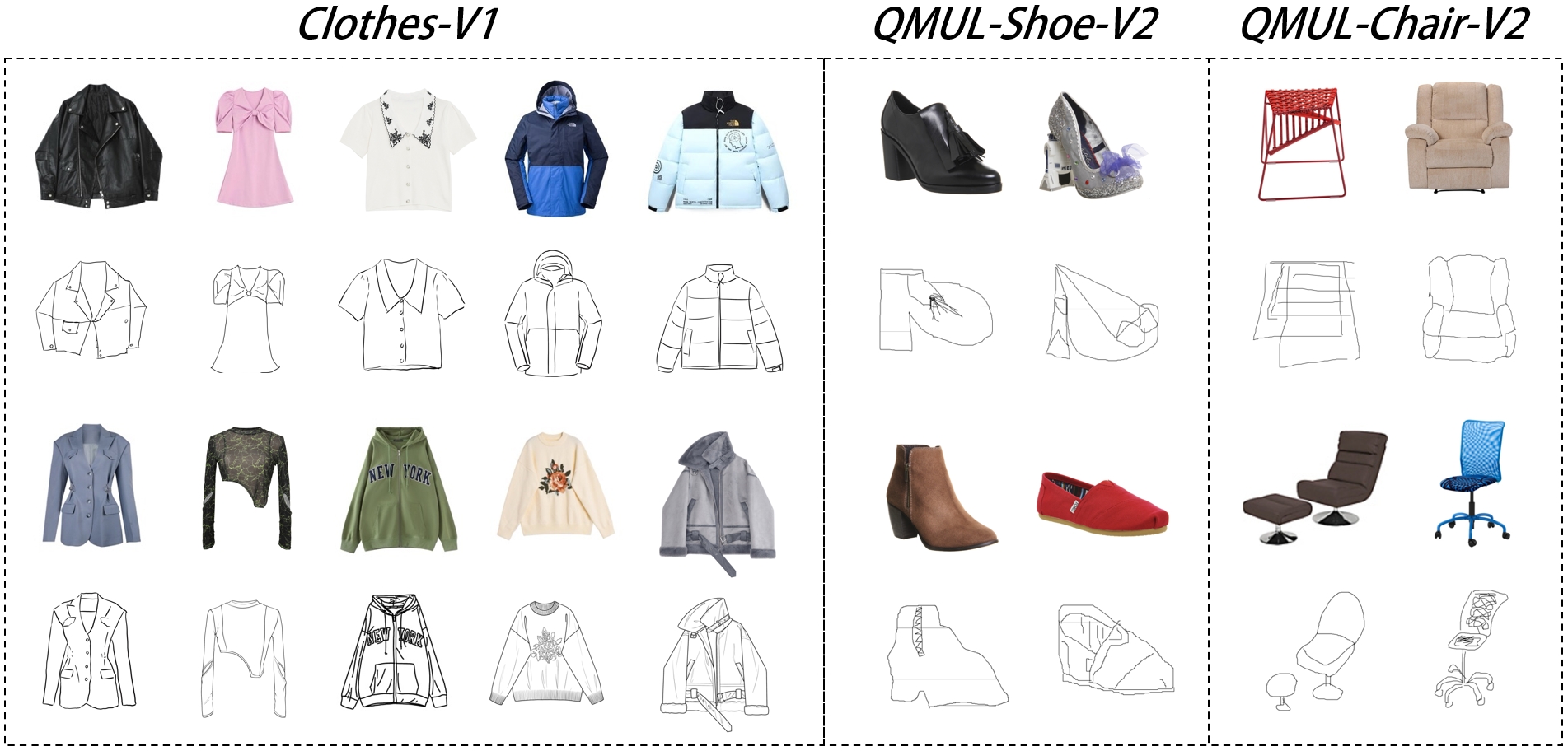}
\caption{Comparison of our dataset with other datasets.}
\label{fig:clothes}
\vspace{-2mm}
\end{figure}

\subsection{Experimental Settings}
\label{section:dataset}
\subsubsection{Clothes-V1.}
Existing fine-grained sketch-based image retrieval (FG-SBIR) datasets~\cite{yu2016sketch} present several challenges: (i) There is a limited number of images without corresponding sketches. (ii) The sketches are drawn in a single rough style. (iii) Fashion clothing, being an essential aspect of our daily lives and a topic of interest for many individuals, lacks a dedicated dataset in the FG-SBIR field for fashion designers. To address these limitations, we invited four professional designers (over five years) and three junior designers (over one year or design novice). This diverse team composition ensures a wide range of sketch styles and levels of expertise, making the dataset more representative of real-life scenarios. \textit{Its specialized nature and diverse collection} of fashion sketches make it a valuable resource for various tasks, including FG-SBIR, image generation, image translation, and fashion-related research. 

Specifically, Clothes-V1 has 1200 (500) sketches (images), containing 925 (380) and 275 (120) for the training and validation set. Each image corresponds to at least one sketch and at most five sketches. The dataset is further categorized into types such as sweaters, shirts, jackets, dresses, cheongsams, etc. Please refer to our GitHub for more details.

\subsubsection{Datasets.}
We utilize two widely used FG-SBIR datasets \textit{QMUL-Chair-V2} and \textit{QMUL-Shoe-V2}~\cite{yu2016sketch} along with our proposed self-collected dataset Clothes-V1 to evaluate the performance of our proposed framework. The \textit{QMUL-Chair-V2} includes 964 (300) and 311 (100) sketches (images) for the training set and validation set, and the \textit{QMUL-Shoe-V2} includes 5982 (1800) and 666 (200) sketches (images) for the training set and validation set.

\begin{table}[t]
\caption{Comparative results of our model against other methods on Clothes-V1.}
\label{tab:Compare_cloth}
\centering
\resizebox{0.76\linewidth}{!}{
\begin{tabular}{lccc}
\toprule
\multirow{2}{*}{Method}
& \multicolumn{3}{c}{Clothes-V1(\%)} \\ \cmidrule(lr){2-4}
& Acc.@1 & Acc.@5 & Acc.@10 \\ \midrule \midrule
Triplet-SN & 64.36 & 85.82 & 92.73 \\
Triplet-Att-SN & 70.18 & 83.64 & 91.64 \\
B-Siamese & 84.73 & 97.82 & \textbf{99.27} \\ 
OnTheFly & 63.27 & 90.18 & 92.73 \\ 
EUPS-SBIR & 89.47 & 94.73 & 97.36 \\ \midrule
\textbf{ARNet Basic} & \textbf{94.12} & \textbf{98.91} & \textbf{99.27} \\
\textbf{ARNet Full} & \textbf{95.27} & \textbf{98.55} & \textbf{99.27} \\ \bottomrule
\end{tabular}
}
\vspace{-2mm}
\end{table}

\subsubsection{Implementation Details.}
Our model is implemented using the PyTorch framework and is based on the ViT-B/16-1K model~\cite{dosovitskiy2020image}. The input size is $224 \times 224$, the final embedding vector's dimension is 512, and the temperature parameter $\tau$ is 0.07. We train the model on a single NVIDIA 32GB Tesla V100 GPU, using a batch size of 16 and the Adam optimizer~\cite{kingma2014adam} with a learning rate of 6e-6 and weight decay of 1e-4, and the training process lasts for 500 epochs. In addition, we incorporated other pre-trained models VGG16~\cite{simonyan2014very}, ResNet50~\cite{he2016deep}, InceptionV3~\cite{szegedy2016rethinking}, and Swin-ViT~\cite{liu2021Swin} to further verify the adaptability and efficacy of our framework.

\subsubsection{Evaluation Metrics.}
We utilize the widely used R@1, R@5, and R@10 to quantify the retrieval performance, which represents the probability of the correct image appearing in the first 1, 5, and 10 retrieved results, respectively.

\subsubsection{Competitors.}
\label{section:com}
We compare our approach with 12 state-of-the-art methods: \textit{Triplet-SN}~\cite{yu2016sketch} adopts triplet loss for model training. \textit{Triplet-Att-SN}~\cite{song2017deep} introduces an attention module. \textit{OnTheFly}~\cite{bhunia2020sketch} employs reinforcement learning and multi-stage sketch input. \textit{B-Siamese}~\cite{sain2020cross} is builds on \textit{Triplet-SN}, but with a stronger backbone. \textit{CMHM-SBIR}~\cite{sain2020cross} proposes a cross-modal multi-level structure. \textit{SketchAA}~\cite{yang2021sketchaa} proposes a multi-granularity modeling approach. \textit{Semi-Sup}~\cite{bhunia2021more} introduces a semi-supervised framework for cross-modal retrieval. \textit{StyleMeUp}~\cite{sain2021stylemeup} proposes a meta-learning framework for unseen sketch styles. \textit{Adpt-SBIR}~\cite{bhunia2022adaptive} introduces a model-agnostic meta-learning framework. \textit{Part-SBIR}~\cite{chowdhury2022partially} proposes cross-modal domain associations by optimal transporting. \textit{NT-SBIR}~\cite{bhunia2022sketching} uses reinforcement learning for sketch stroke subset selection. \textit{EUPS-SBIR}~\cite{sain2023exploiting} employs triplet loss and pre-trained teacher model.


\begin{table}[t]
\caption{Comparative results of our model against other methods on QMUL-Chair-V2 and QMUL-Shoe-V2.}
\label{tab:Compare}
\centering
\resizebox{\linewidth}{!}{
\begin{tabular}{lcccccc}
\toprule
\multirow{2}{*}{Method}
& \multicolumn{3}{c}{QMUL-Chair-V2} & \multicolumn{3}{c}{QMUL-Shoe-V2} \\ \cmidrule(lr){2-4} \cmidrule(lr){5-7} 
& R@1 & R@5 & R@10 & R@1 & R@5 & R@10 \\ \midrule \midrule
Triplet-SN & 33.75 & 65.94 & 79.26 & 18.62 & 43.09 & 59.31 \\
Triplet-Att-SN & 37.15 & 67.80 & 82.97 & 22.67 & 51.20 & 65.02 \\
B-Siamese & 40.56 & 71.83 & 85.76 & 20.12 & 48.95 & 63.81 \\ 
CMHM-SBIR & 51.70 & 80.50 & 88.85 & 29.28 & 59.76 & 74.62 \\
OnTheFly & 39.01 & 75.85 & 87.00 & 35.91 & 66.78 & 78.54 \\
SketchAA & 52.89 & - & 94.88 & 32.22 & - & 79.63 \\
Semi-Sup & 60.20 & 78.10 & 90.81 & 39.10 & 69.90 & \textbf{87.50} \\
StyleMeUp & 62.86 & 79.60 & 91.14 & 36.47 & 68.10 & 81.83 \\
Adpt-SBIR & - & - & - & 38.30 & \textbf{76.60} & - \\
Part-SBIR & 63.30 & 79.70 & - & 39.90 & 68.20 & 82.90 \\
NT-SBIR & 64.80 & 79.10 & - & 43.70 & 74.90 & - \\
EUPS-SBIR & 71.22 & 80.10 & 92.18 & \textbf{44.18} & 70.80 & 84.68 \\  
\midrule
\textbf{ARNet Basic} & \textbf{73.31} & \textbf{93.24} & \textbf{97.15} & 40.11 & 67.54 & 79.29 \\
\textbf{ARNet Full} & \textbf{75.45} & \textbf{94.66} & \textbf{97.87}  & 42.91 & 72.95 & 81.72 \\ \bottomrule
\end{tabular}
}
\end{table}

\subsection{Performance Analysis}
\subsubsection{FG-SBIR.}
As presented in Tables~\ref{tab:Compare_cloth} and~\ref{tab:Compare}, our proposed method exhibits superior performance compared to other baselines on three datasets. 
\textbf{(i)} $\mathtt{QMUL\text{-}Chair\text{-}V2}$: Our model achieves new SOTA results and its Top5 accuracy can surpass EUPS-SBIR's Top10 accuracy. For Top1 accuracy, it outperforms nearly 10\% compared to NT-SBIR and Part-SBIR. \textbf{(ii)} $\mathtt{QMUL\text{-}Shoe\text{-}V2}$: Although our model does not secure the top ranking, it still demonstrates encouraging potential with third position, boasting a slight gap in three metrics compared to the existing methods. This can be attributed to the dataset's larger size and the more relatively irregular sketch strokes compared to the chair and clothes, leading to more complex mappings and being closely clustered within the joint embedding space, which may affect the optimization process. \textbf{(iii)} $\mathtt{Clothes\text{-}V1}$: The results present a clear advantage of our model in Top1 and Top5 accuracy, along with an impressive nearly 100\% accuracy in Top10. \textbf{(iv)} $\mathtt{ARNet~Full}$: Based on the ARNet Basic, our model achieved excellent results on three datasets with steady improvement by our proposed \textit{MSTR Module}. These experimental results demonstrate our model's remarkable retrieval capabilities in a straightforward way, as well as the effectiveness of our approach for recycling discarded Patch Tokens.

\begin{figure}[t]
\centering
\includegraphics[width=\linewidth]{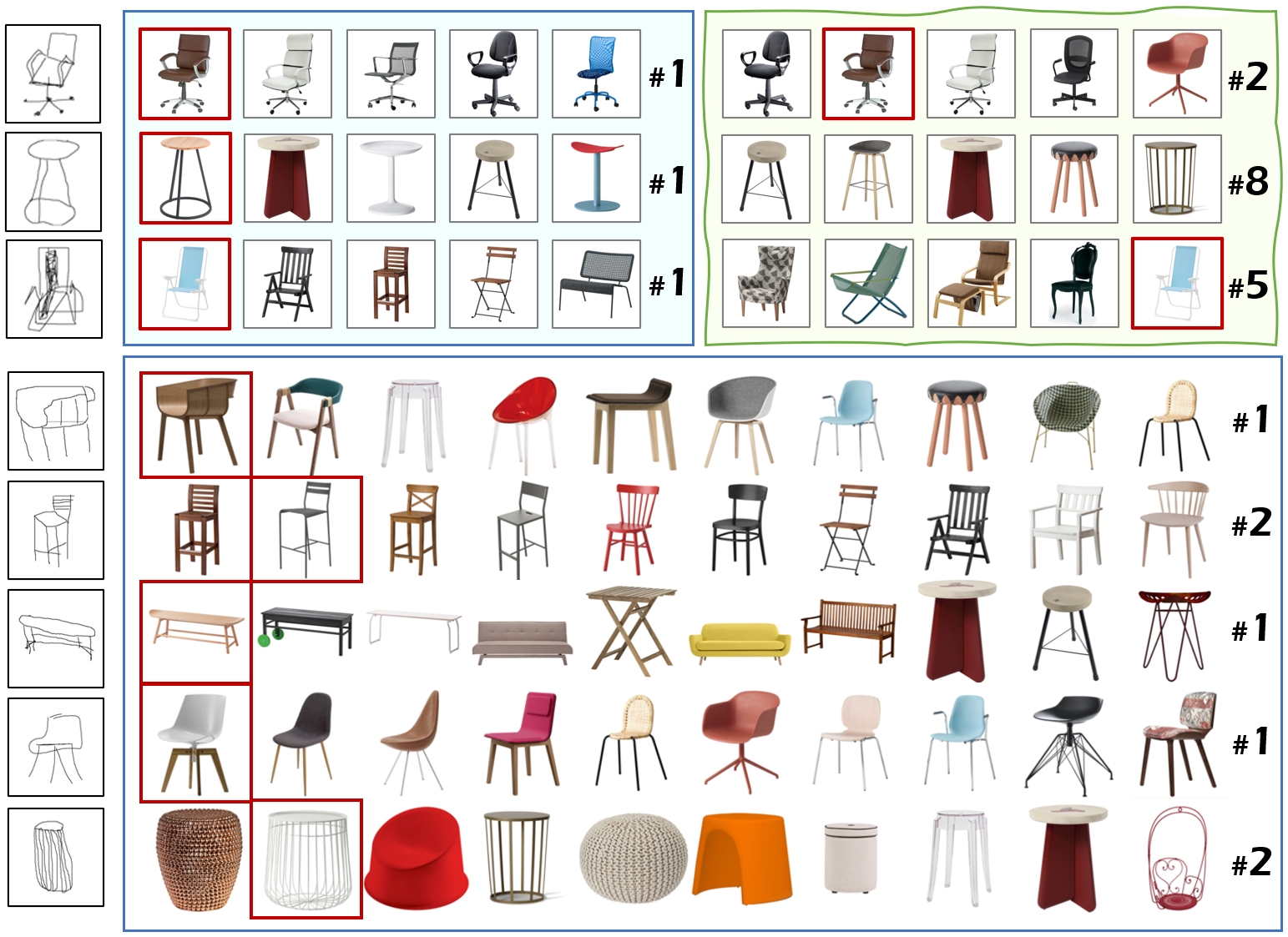}
\caption{Retrieval results between our method (blue) and Triplet-Att-SN (green) on QMUL-Chair-V2.}
\label{fig:retrieval}
\end{figure}

\begin{figure}[t]
\centering
\includegraphics[width=\linewidth]{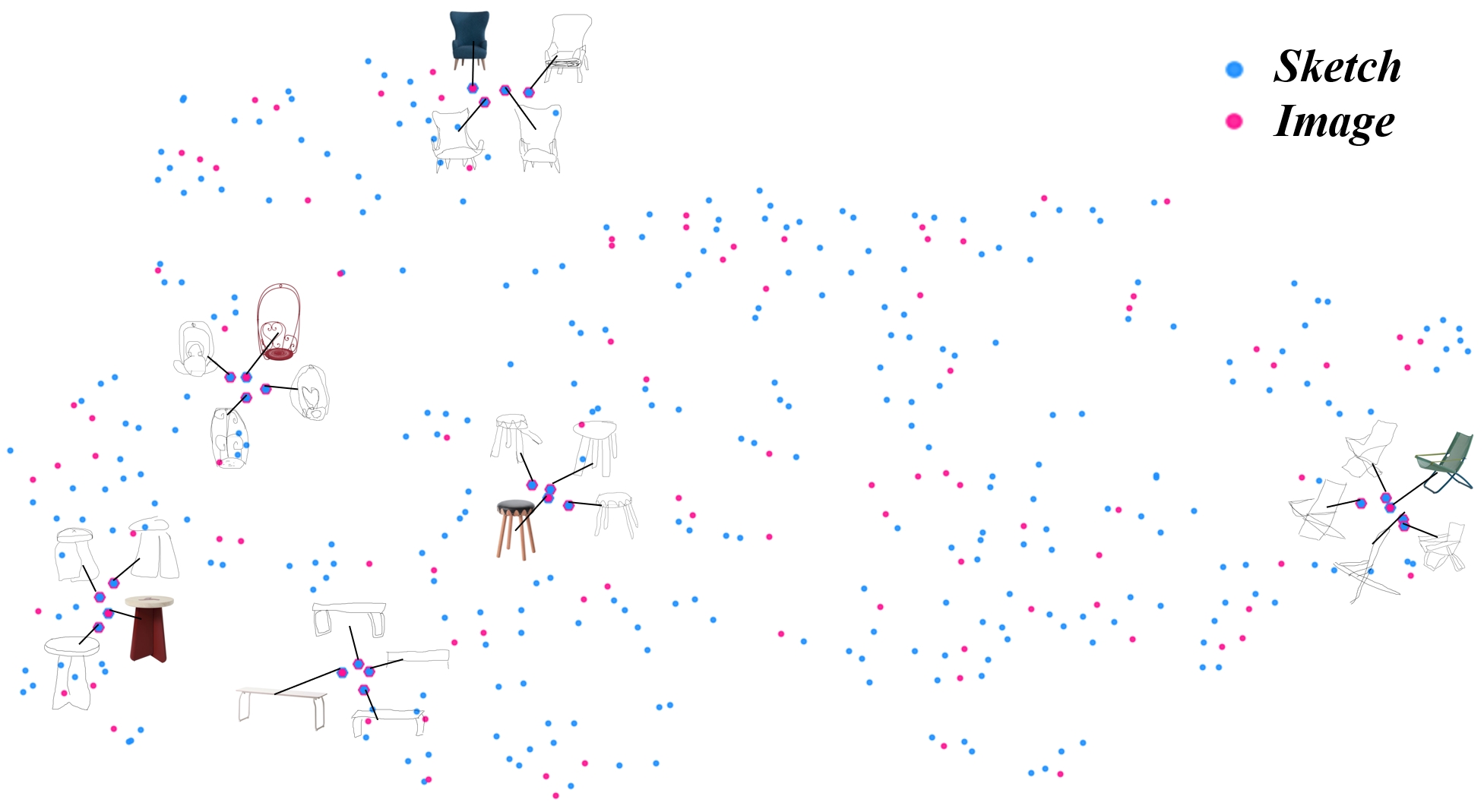}
\caption{t-SNE visualizations of sketch feature (blue) and image feature (pink) embeddings in a joint embedding space on QMUL-Chair-V2 (Zoom in for better visualization).}
\label{fig:tsne}
\end{figure}

\subsubsection{Retrieval Performance.}
We present the comparative retrieval performance in Fig.~\ref{fig:retrieval}. It is evident that the corresponding images retrieved by our method achieve better ranking results in Acc.@5 (top). The top-retrieved image in the Acc.@10 also exhibits a significant similarity to the provided sketch (bottom). Furthermore, in the second row, despite the ground truth being in second place, the first one bears a striking resemblance to the provided sketch from a human visual perspective. These results highlight our stronger retrieval performance.

\subsubsection{Feature Embedding.}
We utilize the t-SNE~\cite{van2008visualizing} to visualize the distribution of encoded sketch and image feature vectors in the joint embedding space. As depicted in Fig.~\ref{fig:tsne}, it illustrates a relatively even distribution across two modalities and a close distribution between images and their corresponding sketches. This indicates the success of our method in extracting and encoding key information, effectively optimizing sample feature distribution within the joint embedding space and consequently yielding better retrieval results.

\subsection{Ablation Study}
\label{section:ab}

\begin{table}[t]
\caption{Ablation Studies on our framework and different ways to load datasets on QMUL-Chair-V2. ((i) $\mathtt{Double}$: Utilized our dual weight-sharing networks. (ii) $\mathtt{Single}$: no utilization. (iii) $\mathtt{Trip\text{-}S/I}$: Use the sketch/image sample as Triplet-loss anchor. (iv) $\mathtt{C/T}$: Contrastive loss/Triplet loss.)}
\label{tab:siam}
\centering
\resizebox{\linewidth}{!}{
\begin{tabular}{cccccccc}
\toprule
\multirow{2}{*}{} & \multirow{2}{*}{ID} & \multirow{2}{*}{Method} & \multirow{2}{*}{Backbone} & \multirow{2}{*}{$\ell_{inter}$} & \multicolumn{3}{c}{QMUL-Chair-V2(\%)} \\ \cmidrule(lr){6-8}
& & & & & Acc.@1 & Acc.@5 & Acc.@10 \\ \midrule \midrule
\multirow{15}{*}{\rotatebox[origin=c]{90}{Framework}}
& 1 & $\mathtt{Single}$ & VGG16 & \tiny \XSolid & 22.42 & 56.23 & 74.38 \\
& 2 & $\mathtt{Single}$ & InceptionV3 & \tiny \XSolid & 39.86 & 78.29 & 90.04 \\
& 3 & $\mathtt{Single}$ & ResNet50 & \tiny \XSolid & 36.66 & 74.73 & 86.48 \\
& 4 & $\mathtt{Single}$ & ViT-B/16 & \tiny \XSolid & 59.43 & 89.32 & 95.02 \\ 
& 5 & $\mathtt{Single}$ & Swin-ViT & \tiny \XSolid & 62.63 & 93.24 & 97.51 \\ \cmidrule(lr){2-8}
& 6 & $\mathtt{Double}$ & VGG16 & \tiny \XSolid & 31.32 & 64.77 & 81.14 \\ 
& 7 & $\mathtt{Double}$ & InceptionV3 & \tiny \XSolid & 42.71 & 80.43 & 90.39 \\
& 8 & $\mathtt{Double}$ & ResNet50 & \tiny \XSolid & 42.35 & 72.95 & 85.41 \\ 
& 9 & $\mathtt{Double}$ & ViT-B/16 & \tiny \XSolid & 58.36 & 90.75 & 95.37 \\ 
& 10 & $\mathtt{Double}$ & Swin-ViT & \tiny \XSolid & 63.35 & 90.75 & 95.37 \\ \cmidrule(lr){2-8}
& 11 & $\mathtt{Double}$ & VGG16 & \checkmark & \textbf{46.26} & \textbf{79.00} & \textbf{86.83} \\ 
& 12 & $\mathtt{Double}$ & InceptionV3 & \checkmark & \textbf{46.26} & \textbf{82.56} & \textbf{91.82} \\
& 13 & $\mathtt{Double}$ & ResNet50 & \checkmark & \textbf{48.04} & \textbf{81.14} & \textbf{89.32} \\ 
& \textbf{14} & $\mathtt{\textbf{Double}}$ & \textbf{ViT-B/16} & \checkmark & \textbf{73.31} & \textbf{93.24} & \textbf{97.15} \\ 
& 15 & $\mathtt{Double}$ & Swin-ViT & \checkmark & \textbf{73.31} & \textbf{93.95} & \textbf{98.22}\\ \bottomrule \toprule
& ID & Method & Augment & Loss & Acc.@1 & Acc.@5 & Acc.@10 \\ \midrule \midrule
\multirow{13}{*}{\rotatebox[origin=c]{90}{Dataloader}}
& 1 & $\mathtt{Single}$ & A0 & $\mathtt{C}$ & 59.43 & 89.32 & 95.02 \\
& 2 & $\mathtt{Single}$ & A1 & $\mathtt{C}$ & 68.33 & 90.39 & 95.73 \\
& 3 & $\mathtt{Single}$ & A2 & $\mathtt{C}$ & 61.92 & 91.04 & 96.44 \\ \cmidrule(lr){2-8}
& \textbf{4} & $\mathtt{\textbf{Double}}$ & \textbf{A0+A1} & $\mathtt{\textbf{C}}$ & \textbf{73.31} & \textbf{93.24} & \textbf{97.15} \\
& 5 & $\mathtt{Double}$ & A0+A2 & $\mathtt{C}$ & 69.75 & 92.17 & 95.37 \\
& 6 & $\mathtt{Double}$ & A1+A2 & $\mathtt{C}$ & 64.41 & 87.19 & 93.24 \\
& 7 & $\mathtt{Double}$ & A2+A1 & $\mathtt{C}$ & 58.36 & 85.41 & 91.82 \\ \cmidrule(lr){2-8}
& 8 & $\mathtt{Trip\text{-}S}$ & A0 & $\mathtt{T}$ & 57.65 & 89.83 & 96.09 \\
& 9 & $\mathtt{Trip\text{-}S}$ & A1 & $\mathtt{T}$ & 55.52 & 87.90 & 95.37 \\
& 10 & $\mathtt{Trip\text{-}S}$ & A2 & $\mathtt{T}$ & 51.60 & 81.85 & 90.75 \\ \cmidrule(lr){2-8}
& 11 & $\mathtt{Trip\text{-}I}$ & A0 & $\mathtt{T}$ & 24.20 & 51.96 & 68.68 \\
& 12 & $\mathtt{Trip\text{-}I}$ & A1 & $\mathtt{T}$ & 16.01 & 41.64 & 55.87 \\
& 13 & $\mathtt{Trip\text{-}I}$ & A2 & $\mathtt{T}$ & 17.44 & 41.28 & 58.36 \\ \bottomrule
\end{tabular}
}
\vspace{-2mm}
\end{table}

\subsubsection{Framework.}
Table~\ref{tab:siam} represents that our dual weight-sharing networks can significantly enhance the overall performance of the model across various mainstream backbone networks. In essence, our framework takes into account both the single feature distribution within the sketches and images while considering the joint feature distribution of the sketches and images, resulting in more representative feature representations. Furthermore, our observations revealed that the inclusion of $\ell_{inter}$ not only enhances the overall performance of the model, but also accelerates the convergence trajectory. This can be attributed to its ability to further enhance the feature alignment between sketches and images based on the model structure. This increased alignment contributes to the extraction of more distinctive and representative features by dual weight-sharing networks.

\subsubsection{Dataloader.}
We compare three different data processing strategies using two data loading methods based on the Contrast loss $\mathtt{C}$ and the Triplet loss $\mathtt{T}$ in Table~\ref{tab:siam}. The results demonstrated that the $\mathtt{C}$-based method consistently outperforms the $\mathtt{T}$-based method under the same conditions. This superiority is attributed to our loss function's comprehensive in-sample and inter-sample optimization, which provides greater gradient contrast information in each mini-batch optimization, thereby enhancing overall performance.

\begin{table}[t]
\caption{Ablation Studies on our MSTR Module on QMUL-Chair-V2. ($\mathtt{max}$, $\mathtt{mean}$, and $\mathtt{cat}$: Use MaxPool, AvgPool, and Concatenation for Discarded Patch Tokens.)}
\label{tab:mstr}
\centering
\resizebox{\linewidth}{!}{
\begin{tabular}{ccccccccccc}
\toprule
\multirow{2}{*}{} & \multirow{2}{*}{ID} & \multirow{2}{*}{Rcyc} & \multirow{2}{*}{Map} & \multirow{2}{*}{LayT} & \multirow{2}{*}{LayS} & \multirow{2}{*}{Dim} & \multicolumn{3}{c}{QMUL-Chair-V2(\%)} \\ \cmidrule(lr){8-10} 
& & & & & & & Acc.@1 & Acc.@5 & Acc.@10 \\ \midrule \midrule
\multirow{11}{*}{\rotatebox[origin=c]{90}{MSTR Module}}
& 1 & \tiny \XSolid & \tiny \XSolid & \tiny \XSolid & \tiny \XSolid & \tiny \XSolid & 73.31 & 93.24 & 97.15 \\
& 2 & $\mathtt{max}$ & \tiny \XSolid & \tiny \XSolid & \tiny \XSolid & \tiny \XSolid & 72.95 & 94.66 & \textbf{98.22} \\
& 3 & $\mathtt{mean}$ & \tiny \XSolid & \tiny \XSolid & \tiny \XSolid & \tiny \XSolid & 71.89 & 93.59 & 97.15 \\
& 4 & $\mathtt{cat}$ & \tiny \XSolid & \tiny \XSolid & \tiny \XSolid & \tiny \XSolid & 65.48 & 92.53 & 97.51 \\
& 5 & \checkmark & \tiny \XSolid & 1 & 3 & 16 & 73.67 & 94.31 & 97.51 \\
& 6 & \checkmark & \checkmark & 1 & 3 & 16 & 74.38 & 94.66 & 97.15 \\
& 7 & \checkmark & \checkmark & 3 & 3 & 16 & 71.89 & 93.24 & 96.44 \\
& \textbf{8} & \textbf{\checkmark} & \textbf{\checkmark} & \textbf{2} & \textbf{3} & \textbf{16} & \textbf{75.45} & \underline{94.66} & \underline{97.87} \\
& 9 & \checkmark & \checkmark & 2 & 4 & 16 & 72.95 & 92.53 & \underline{97.87} \\ 
& 10 & \checkmark & \checkmark & 2 & 2 & 16 & 73.67 & \textbf{95.02} & 95.02 \\
& 11 & \checkmark & \checkmark & 2 & 3 & 8 & 72.60 & 93.59 & 97.15 \\
& 12 & \checkmark & \checkmark & 2 & 3 & 24 & 73.31 & 94.31 & 97.51 \\ 
& 13 & \checkmark & \checkmark & 2 & 3 & 48 & 72.60 & 93.59 & 97.15 \\
\bottomrule
\end{tabular}
}
\vspace{-2mm}
\end{table}

\subsubsection{MSTR Module.}
As shown in Table~\ref{tab:mstr}, we conducted ablation experiments on the MSTR module from four perspectives: recycling discarded Patch Tokens (Rcyc), constructing $\mathtt{MAP}$ (Map), the number of Transformer Layers (LayT), the number of feature Scale Layers (LayS), and the feature dimension used for Map (Dim). The results indicate that simple methods like MaxPool, AvgPool, and Concatenation not only fail to improve but may even negatively impact model performance. Using the map can enhance contrastive learning between tokens, thus improving model performance. When examining LayT, 2 layers outperformed both 1 and 3 layers. This is mainly because 1 layer fails to encode feature information adequately, while 3 layers result in over-encoding and negative effects. For validating LayS, we evaluated models with 2 and 4 layers compared to our 3-layer setting. The results demonstrate that using 3 layers improves Top1 accuracy by approximately 2\% and 3\% compared to the settings of 2 and 4 layers, while also achieving balanced performance in Top5 and Top10 accuracy. Regarding Dim, we observed a similar phenomenon that 8 cannot encapsulate all unique information, while 24 and 48 include too much irrelevant information, weakening the feature representation of discarded \textit{Patch Tokens}. These findings collectively demonstrate the effectiveness of our MSTR module.

\section{Conclusion}
In this paper, we present ARNet, a novel FG-SBIR solution, along with a general and diverse clothing dataset, Cloth-V1. Our approach incorporates structure and loss function design, enabling the model to concurrently focus on intra- and inter-sample feature alignment. Our proposed method exhibits strong scalability, accommodating various mainstream backbones as feature extraction networks, and offers straightforward implementation. Moreover, we introduce the MSTR module to further improve our basic model representation ability by recycling discarded Patch Tokens. Extensive experiments demonstrate that our ARNet achieves superior retrieval performance across several benchmarks.

                           
\bibliography{main}

\end{document}